\documentclass[11pt,a4paper]{article}
\usepackage{authblk}
\usepackage[hyperref]{acl2020}
\usepackage{times}
\usepackage{latexsym}
\usepackage{graphicx}
\usepackage{amsfonts}
\usepackage{amsmath}
\usepackage{dsfont}
\usepackage{multirow}
\usepackage[ruled,vlined]{algorithm2e}
\usepackage{mathtools}

\usepackage{placeins}

\usepackage{microtype}

\aclfinalcopy 

\title{Assertion Detection in Multi-Label Clinical Text using Scope Localization}

\author[1]{Rajeev Bhatt Ambati}
\author[1,2]{Ahmed Ada Hanifi}
\author[1]{Ramya Vunikili}
\author[1]{Puneet Sharma}
\author[1]{Oladimeji Farri}
\affil[1]{Siemens Healthineers, Princeton, NJ, USA \protect\\ \texttt{\{rajeevbhatt.ambati, ramya.vunikili,} \protect\\ \texttt{sharma.puneet, oladimeji.farri\}@siemens-healthineers.com}}
\date{}
\affil[2]{\texttt{ahmed.ada.aa@gmail.com}}
\begin{document}
\maketitle
\begin{abstract}

Multi-label sentences (text) in the clinical domain result from the rich description of scenarios during patient care. The state-of-the-art methods for assertion detection mostly address this task in the setting of a single assertion label per sentence (text). In addition, few rules based and deep learning methods perform negation/assertion scope detection on single-label text. It is a significant challenge extending these methods to address multi-label sentences without diminishing performance. Therefore, we developed a convolutional neural network (CNN) architecture to localize multiple labels and their scopes in a single stage end-to-end fashion\footnote{Disclaimer: The concepts and information presented in this paper are based on research results that are not commercially available}, and demonstrate that our model performs atleast 12\% better than the state-of-the-art on multi-label clinical text.
\end{abstract}

\section{Introduction}
\par In recent years, advanced natural language processing (NLP) techniques have been applied to electronic health record (EHR) documents to extract useful information. Accessibility to large scale EHR data is very crucial to using such deep learning methods - yet data scarcity persists for most tasks in the healthcare domain. 
\par Assertion detection involves classifying clinical text obtained from the EHR and other hospital information systems (e.g. Radiology Information System/RIS), to determine if a medical concept (entity) is \textit{present}, \textit{absent}, \textit{conditional}, \textit{hypothetical}, \textit{possibility} or \textit{AWSE} (associated with someone else). These classes were used in \citet{Chen2019AttentionBasedDL}. A few examples of each class from our dataset are shown in Table.~\ref{tab:class_examples}.

\begin{figure}[ht!]
    \centering
    \includegraphics[width=0.8\linewidth]{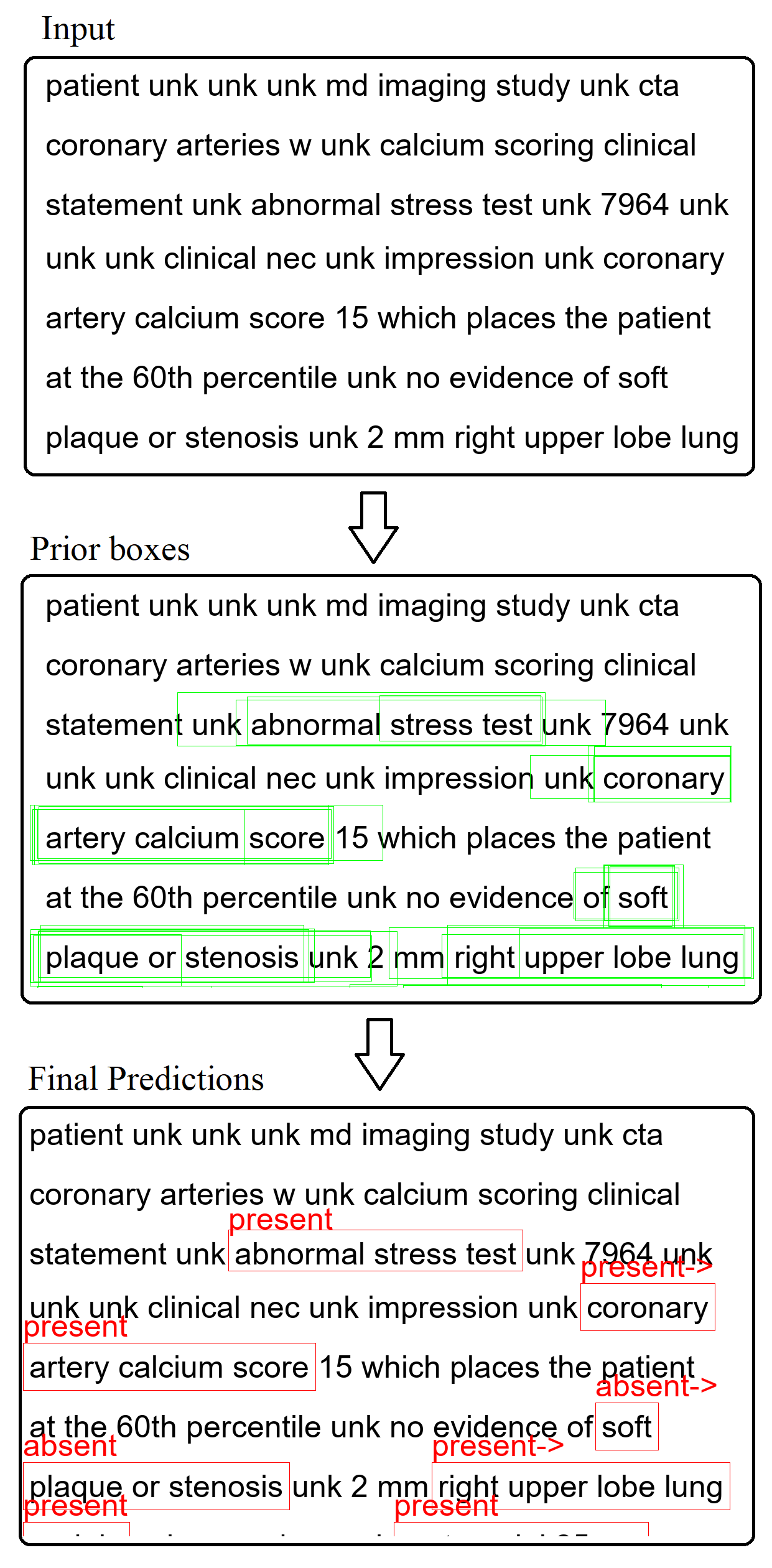}
    \caption{From the Input, the model will predict confidence scores for all the prior boxes at each token. Prior boxes with confidence threshold greater than $\gamma$ are shown in green. After non-max suppression, the final predictions are shown in red.}
    \label{fig:boxesintro}
\end{figure}

\par Past works with the i2b2 dataset mostly focused on the \textit{present} and \textit{absent} classes with comparatively less work on the more ambiguous classes. Majority of the existing methods either classify the given text only, or use the class further to detect it's scope in a two stage process. This works well for datasets like i2b2 \citep{i2b2challenge} in which there exists only one label per example. However, single label per sentence is not a common phenomenon in clinical reports, especially when patients have frequent physician visits or long periods of hospitalization. To address the aforementioned problem, our work highlights the following contributions:
\begin{itemize}
    \item We explored assertion detection in multi-label sentences from radiology (cardiac computerized tomography (CT)) reports.
    \item We cast the assertion detection task as a \textit{scope localization} problem, thereby solving classification and scope detection in a single stage end-to-end fashion.
    \item We leveraged concepts from object localization \cite{YOLO} in computer vision and developed a CNN to detect bounding boxes around class scopes.
\end{itemize}

\begin{figure}
    \centering
    \includegraphics[width=0.8\linewidth]{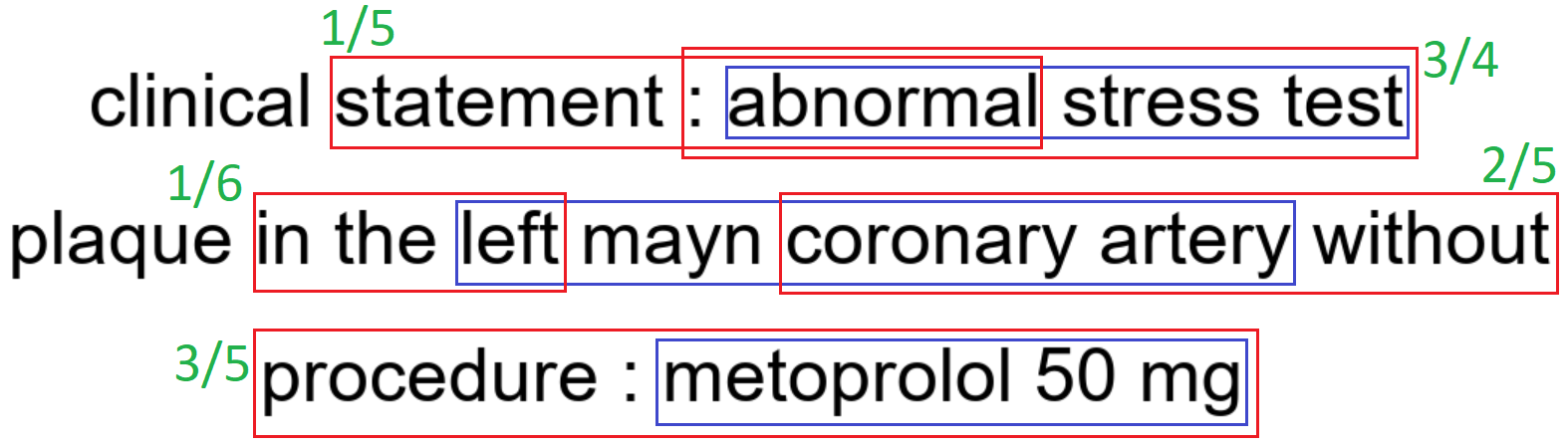}
    \caption{The blue boxes are ground-truths. The IoU of each of the red boxes with the blue boxes are shown in green.}
    \label{fig:iou_example}
\end{figure}

\section{Related Work}
\par Rule based models like NegEx\cite{negex}, NegBio \cite{negbio} and \cite{gkotsis-etal-2016-dont} were initially used for assertion and negation detection. These approaches typically implement rules and regular expressions to detect cues for classification. NegBio \cite{negbio} uses a universal dependency graph to detect the scope of identified class. A constituency parsed tree is used by \citet{gkotsis-etal-2016-dont} to prune out words outside the scope of the detected class. NegEx \cite{negex} later demonstrated good performance when adapted to many other languages like German, Spanish and French \cite{cotik_et_al_2015, stricker_et_al_2015, costumero2014approach, Afzal2014ContextDAA}. A few approaches developed syntactic techniques by augmenting dependency parsed trees to rule based systems \cite{Mehrabi2015DEEPENAN, Sohn2012DependencyPN, cotik-etal-2016-syntactic}. \citet{Mackinlay_et_al_2012} constructed hand-engineered features using English Resource Grammar to identify negation and hypothetical classes for a BIONLP 2009 task.

\begin{table}
\centering
    \begin{tabular}{ p {2cm} p {5cm} }
        \hline
        \textbf{Class} & \textbf{Examples} \\
        \hline
        Present & \textbf{Metoprolol  50 mg} po was administered prior to the scan to decrease heart rate \\
        Absent & No \textbf{Chest pain}, No \textbf{Coronary artery Aneurysm}, No \textbf{Aneurysm or wall thickening} \\
        Conditional & \textbf{Myocardial perfusion imaging}, if not already performed, might improve specificity in this regard if clinically warranted \\
        Hypothetical & \textbf{Coronary plaque burden and focal Lesion} characterization (if present) were assessed by visual estimate. \\
        Possibility & This was incompletely imaged but suggests a \textbf{diaphragmatic arteriovenous malformation} \\
        AWSE & High risk is $>$ or = 10 packs/year or positive family history of \textbf{lung cancer} in first degree relative \\
        \hline
    \end{tabular}
\caption{Examples of each class from our dataset. From each sentence (or phrase) shown, the text in bold is identified as the corresponding class.}
\label{tab:class_examples}
\end{table}

\par The annotated entities and  assertion /labels in the 2010 i2b2/VA challenge \citep{i2b2challenge} can be regarded as a benchmark for the assertion detection task for clinical text. Kernel methods using SVM \citep{Bruijn2011MachinelearnedSF} and Bag-of-Words \citep{shivade-etal-2015-extending} were proposed for the shared task. \citet{cheng-etal-2017-automatic} used a CRF for classification of cues and scope detection. Though these methods have performed better than rule based methods, they fail to generalize well to unseen examples while training.

\par More recently, with the advent of deep learning achieving state-of-the-art performance in various NLP tasks, an LSTM encoder-decoder architecture \citep{DBLP:journals/corr/SutskeverVL14} \citep{Hochreiter:1997:LSM:1246443.1246450} can be trained for assertion detection with reasonable success. Attention based models using LSTMs \citep{fancellu-etal-2016-neural} and GRUs \citep{Rumeng2017AHN} were explored. Limited amounts of labeled (and unlabeled) clinical text make training deep neural networks a challenging task. \citet{Bhatia-et-al-2018} explored a multi-task learning setting by combining a Named Entity Recognition (NER) classification branch to the assertion detection output. All of these methods either identify only the class or use it as a cue to prune the scope of the class from the text. As mentioned above, our work proposes an end-to-end single stage approach to assertion and negation scope detection. A schematic of our approach is shown in Fig.\ref{fig:boxesintro}.

\section{Proposed Model}
\par We formulated the assertion and negation problem as follows: Let $R = \{r_{1}, r_{2}, ..., r_{T}\}$ be a sentence in clinical report consisting of $T$ words $r_{i}$. We need to identify the $L$ assertion classes and corresponding scope in the report defined by the set $S = \{(c_{1}, x_{1}, y_{1}), (c_{2}, x_{2}, y_{2}), ..., (c_{L}, x_{L}, y_{L})\}$ where, class $c_{i}$ scopes between $x_{i} \in [1, T]$ and $y_{i} \in [1, T]$. We put forward this problem as finding bounding boxes over the text that scope a particular class. If $A$ is the maximum scope of a class present in the input, we can place prior boxes of lengths $\{1, 2, .., A\}$ at each word $r_{i}$ and predict the probability of a particular box containing a class.

\subsection{Intersection Over Union}
Let $B_{1}, B_{2}$ be two bounding boxes over text scopes $T_{1}, T_{2}$ where, $T_{i}$ is a set of words. We then define the IoU (Intersection over Union) of these two bounding boxes as follows:

\begin{equation}
    IoU(B_{1}, B_{2}) = \frac{|T_{1} \cap T_{2}|}{|T_{1} \cup T_{2}|}
\end{equation}

Where $|S|$ is the cardinal of a set $S$. A few examples if IoUs are shown in Fig.\ref{fig:iou_example}.

\subsection{Network Design}
First, we embed the input sequence in a distributional word vector space as $W = \{e_{1}, e_{2}, ..., e_{T}\}$ where, $e_{i}\in \mathbb{R}^{D}$ is a column vector in an embedding matrix $E \in R^{T \times D}$. This is the input to our CNN. Each layer in the CNN is a 1D-convolutional layer followed by a non-linearity. Stacking many layers on top of the other increases the receptive field of the network. To cover the largest prior box of length $A$, we need the receptive field of the last layer to be at least $A$.


\begin{figure*}[htb]
    \centering
    \includegraphics[width=\linewidth]{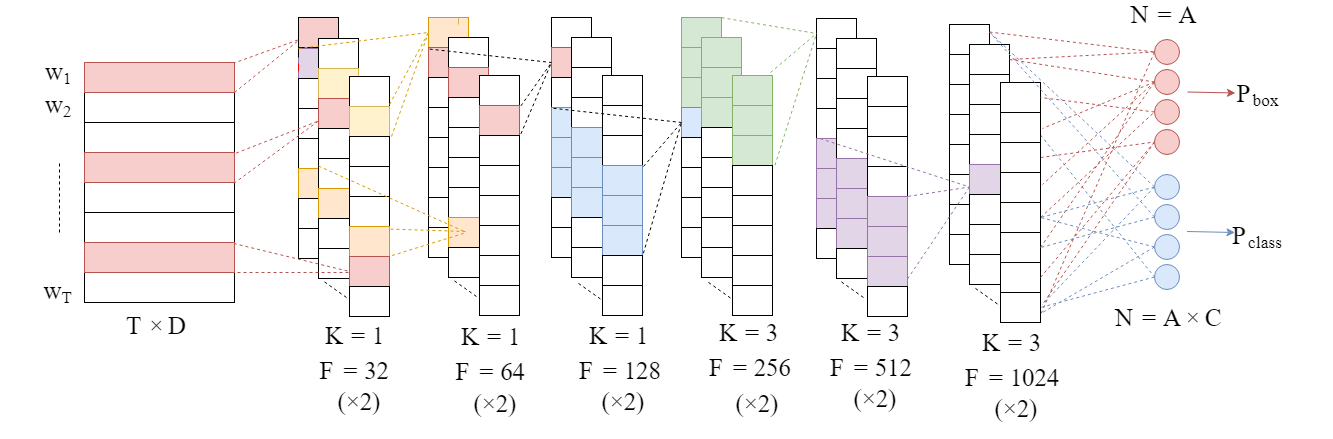}
    \caption{CNN Architecture: The input sequence is first passed through a pretrained embedding layer. $\times 2$ denotes two layers of same kernel size and no. of filters. First 6 layers use a kernels of size 1 and last 6 layers use kernels of size 3. The no. of filters double every two layers. Here, K, F and N are kernel size, filter size and number of units respectively. The feature map of the last convolutional layer is projected using two fully-connected layers of units $A$ and $A \times C$ respectively. The resulting box confidence and class confidence probabilities are fed to the MSE and cross-entropy loss functions respectively.}
    \label{fig:my_label}
\end{figure*}

Our architecture is shown in Fig.\ref{fig:my_label}. First we use 6 layers of $1 \times 1$ convolutions followed by 6 layers of $ 3 \times 1$ convolutions. We use a stride of 1 and pad the feature maps wherever necessary to maintain constant feature map size of $T$ throughout the network. We also use ReLU non-lineartiy after each convolutional layer. The output of the last convolutional layer is then passed through 2 branches of fully-connected layers to produce box confidence scores $p_{box} \in \mathbb{R}^{T \times A}$ and class confidence probabilities $p_{class} \in \mathbb{R}^{T \times A \times C}$ respectively. Where, $A$ is the number of prior boxes and $C$ is the number of classes. It is important to note the receptive field of the last layer is 24.

\subsection{Objective Function}
\textbf{Box Confidence Loss:} We expect the box confidence branch to predict the IoU of each prior box with the nearest ground truth box. The simplest way to do this is by minimizing the Mean Square Error (MSE) between predicted and ground-truth IoU.

\begin{equation}
    L_{box} = \frac{1}{T \times A}\sum_{t=1}^{T}\sum_{a=1}^{A}\lVert p_{box}(t,a) - iou \rVert _{2}^{2}
\end{equation}

\textbf{Non-max Suppression} Once we have the box confidence scores of $T*A$ prior boxes, we sort them in the decreasing order of their confidence scores and discard the ones lower than a confidence threshold $\gamma$. In the remaining overlapping boxes, we vote for the prior box with high confidence score. The detailed algorithm is shown in Algorithm \ref{algo:nms}.

\textbf{Class Confidence Loss:} The class confidence branch is expected to predict $P(class | box)$, the probability of a class given that a prior box has an assertion scope. We first apply softmax on the class confidence score and use cross-entropy loss to maximize the probability of the ground-truth class. Given the class imbalance in the dataset we used, a weighted loss per class was implemented.

\begin{equation}
    L_{class} = \frac{1}{T \times A}\sum_{t=1}^{T}\sum_{a=1}^{A}\mathds{1}_{box}^{a}\sum_{c=1}^{C}-w_{c}\mathds{1}_{c}\log(p_{class}(c))
\end{equation}

Where, $\mathds{1}_{box}^{a}$ is an indicator variable denoting the presence of a class in prior box-$a$ and $w_{c}$ is the weight of class-$c$ which is equal to the fraction of examples in a batch that belong to class-$c$. 
\par We optimize the cummulative loss $L = L_{box} + L_{class}$ using Adam optimizer.


\begin{algorithm}[ht!]
\SetAlgoLined
\KwResult{Final non-overlapping boxes}
$indices = ArgSort(scores)$ \;
$boxes=[ ]$ \;
$i = 0$ \;
 \While{$i \leq T*A$}{
  \eIf{$ scores(i) \leq \gamma $}{
   break\;
   }{
   $maxIoU \gets max(IoU(priors[i], boxes))$ \;
   \If{maxIoU == 0}{
   $boxes \gets [boxes, priors[i]]$ \;
   }
  }
  $i \gets i + 1$ \;
 }
 \caption{NonMaxSup(scores, priors)}
 \label{algo:nms}
\end{algorithm}

\section{Datasets and Experiments}
We evaluated our model on datasets from two hospital sites (Dataset-I and Dataset-II); both have reports with multi-label sentences. First we will elaborate on the data collection and annotation process. Next, we will present some statistics on the datasets and, finally, highlight the performance of our model.
Dataset-I and II comprise 151 and 460 cardiac CT reports respectively. All reports were anonymized at the hospital site before we accessed the data. The datasets were annotated by 8 engineers with an average of 217 hours of training in labeling healthcare data.

The annotations were done using BRAT tool \citep{Stenetorp:2012:BWT:2380921.2380942}. Rules for annotation were generated after consulting with the Radiologist supervising the annotators. Other Radiologists were consulted to annotate any mentions that were previously unseen or ambiguous and also for the final review. Statistics of the data such as No. of classes per report, No. of tokens in a report and length of class scopes are shown in Tables.\ref{tab:split_class_counts_per_class_pr}-\ref{tab:split_tag_lens_per_class}.


\begin{table*}
    \centering
    \begin{tabular}{ c  c  c  c  c  c  c}
        \hline
        \multirow{2}{*}{\textbf{Class}} & \multicolumn{3}{c}{\textbf{Dataset-I}} & \multicolumn{3}{c}{\textbf{Dataset-II}}\\
        & \textbf{Train} & \textbf{Val} & \textbf{Test} & \textbf{Train} & \textbf{Val} & \textbf{Test} \\
        \hline
        Present & 3711 & 511 & 524 & 17407 & 2215 & 2452\\
        Absent & 596 & 73 & 73 & 6136 & 708 & 805\\
        Conditional & 169 & 31 & 19 & 393 & 44 & 49\\
        Hypothetical & 147 & 22 & 18 & 69 & 10 & 5\\
        Possibility & 62 & 5 & 11 & 219 & 37 & 25\\
        AWSE & 15 & 3 & 2 & 21 & 4 & 2\\
        \hline
    \end{tabular}
    \caption{Distribution of Assertion classes in the data.}
    \label{tab:split_class_counts_per_class_pr}
    \vspace*{0.5cm}
    \centering
    \begin{tabular}{ c  c  c  c  c  c  c }
        \hline
        \multirow{2}{*}{\textbf{Split}} & \multicolumn{3}{c}{\textbf{Dataset-I}} & \multicolumn{3}{c}{\textbf{Dataset-II}} \\
         & \textbf{Max} & \textbf{Min} & \textbf{Mean} & \textbf{Max} & \textbf{Min} & \textbf{Mean} \\
        \hline
        train & 661 & 19 & 440 & 1028 & 82 & 610\\
        val & 642 & 289 & 452 & 911 & 82 & 630\\
        test & 560 & 228 & 432 & 968 & 336 & 642\\
        \hline
    \end{tabular}
    \caption{Number of tokens per report in the data.}
    \label{tab:split_rep_lengths_per_class_pr}
    \vspace*{0.5cm}
    \centering
    \begin{tabular}{ c  c  c  c  c  c  c }
        \hline
        \multirow{2}{*}{\textbf{Class}} & \multicolumn{3}{ c }{\textbf{Dataset-I}} & \multicolumn{3}{ c }{\textbf{Dataset-II}}\\
        & \textbf{Train} & \textbf{Val} & \textbf{Test} & \textbf{Train} & \textbf{Val} & \textbf{Test} \\
        \hline
        1 & $3.36 \pm 2.7$ & $3.23 \pm 2.59$ & $3.39 \pm 2.95$ & $3.48 \pm 2.15$ & $3.38 \pm 2.06$ & $3.48 \pm 2.16$\\
        2 & $2.79 \pm 1.36$ & $2.68 \pm 1.25$ & $2.68 \pm 1.07$ & $3.15 \pm 2.26$ & $3.10 \pm 2.18$ & $3.09 \pm 2.13$ \\
        3 & $2.85 \pm 1.04$ & $2.87 \pm 0.87$ & $2.68 \pm 0.65$ & $3.24 \pm 2.92$ & $3.20 \pm 1.95$ & $2.60 \pm 1.74$  \\
        4 & $5.05 \pm 2.44$ & $4.5 \pm 2.44$ & $5.0 \pm 2.43$ & $2.19 \pm 1.21$ & $2.60 \pm 0.92$ & $2.84 \pm 2.47$ \\
        5 & $3.14 \pm 3.69$ & $1.67 \pm 0.47$ & $3.27 \pm 2.41$ & $2.96 \pm 2.83$ & $2.40 \pm 1.82$ & $3.27 \pm 2.41$\\
        6 & $2.47 \pm 0.72$ & $1.67 \pm 0.47$ & $5.5 \pm 2.5$ & $1.71 \pm 1.35$ & $2.00 \pm 1.73$ & $1.0 \pm 0.0$\\
        \hline
    \end{tabular}
    \caption{Scope lengths of each class per train, validation (val) and test splits. Following are the classes corresponding to the IDs 1: \textit{Present}, 2: \textit{Absent}, 3: \textit{Conditional}, 4: \textit{Hypothetical}, 5: \textit{Possibility}, 6: \textit{AWSE} and \textit{macro} $F_{1}$ score is the average over all classes. Lengths are written in the format $\mu \pm \sigma$.}
    \label{tab:split_tag_lens_per_class}
    \vspace*{0.5cm}
    \centering
    \begin{tabular}{ c | c | c | c | c }
        \hline
        \multirow{4}{*}{\textbf{Class}} & \multicolumn{4}{ c }{\textbf{Model}} \\
        \cline{2-5}
          & \multicolumn{2}{ c |}{\textbf{Baseline}} & \multicolumn{2}{ | c  }{\textbf{Scope Localization model}} \\
          \cline{2-5}
          \cline{2-5}
          & \textbf{Dataset-I} & \textbf{Dataset-II} & \textbf{Dataset-I} & \textbf{Dataset-II} \\
          \hline
         Present & 0.97 & 0.92 & 0.90 & 0.84 \\
         Absent & 0.27 & 0.34 & 0.84 & 0.93 \\
         Conditional & 0.39 & 0.45 & 0.74 & 0.65 \\
         Hypothetical & 0.76 & 0.69 & 0.87 & 0.75 \\
         Possibility & 0.0 & 0.07 & 0.0 & 0.13 \\
         AWSE & 0.42 & 0.39 & 0.60 & 0.0 \\
         None & 0.81 & 0.89 & 0.96 & 0.95 \\
         \hline
         Macro & 0.52 & 0.53 & 0.70 & 0.61 \\
         \hline
    \end{tabular}
\caption{The performance of both baseline and our CNN model on Dataset-I and Dataset-II in terms of $F_{1}$ score. \textit{macro} $F_{1}$ score is the average over all classes.}
\label{tab:results}
\end{table*}

\subsection{Baseline Model}

\citep{Bhatia-et-al-2018, Chen2019AttentionBasedDL, Rumeng2017AHN}. \citet{Chen2019AttentionBasedDL} used a bidirectional attentive encoder on the sentence input to obtain a context vector which is subsequently passed to the softmax and output classification layers. \citet{Bhatia-et-al-2018} extended this network by adding a shared decoder to predict both assertion class and named entity tag in a multi-task learning framework. However, the input to these seq2seq models is a sentence and the output prediction is a single class. Therefore, the models may not be easily extended to a multi-label dataset without compromising performance. To validate our assumption, we extend the bidirectional encoder and attentive decoder model based on LSTM to our multi-label data by changing the input format. In other words, instead of predicting one class for the entire input sequence, we predict a class for each token so that the scope of a class can also be localized. Two sample sentences (with class labels) are shown in Table.\ref{tab:baseline_data_format}.

\begin{table}
\centering
    \begin{tabular}{| p {1.2cm} | p {5cm} |}
        \hline
        Report-1 & Metoprolol\textsuperscript{P} 50\textsuperscript{P} mg\textsuperscript{P} po\textsuperscript{N} was\textsuperscript{N} administered\textsuperscript{N} prior\textsuperscript{N} to\textsuperscript{N} the\textsuperscript{N} scan\textsuperscript{N} to\textsuperscript{N} decrease\textsuperscript{C} heart\textsuperscript{C} rate\textsuperscript{C} \\
         \hline
         Report-2 & Myocardial\textsuperscript{H} perfusion\textsuperscript{H} imaging\textsuperscript{H} ,\textsuperscript{N} if\textsuperscript{N} not\textsuperscript{N} already\textsuperscript{N} performed\textsuperscript{N} ,\textsuperscript{N} might\textsuperscript{H} improve\textsuperscript{H} specificity\textsuperscript{H} in\textsuperscript{N} this\textsuperscript{N} regard\textsuperscript{N} if\textsuperscript{N} clinically\textsuperscript{N} warranted\textsuperscript{N} .\textsuperscript{N} \\
         \hline
    \end{tabular}
\caption{Two sample sentences with the label format for the baseline seq2seq model. P, C, H, N denote \textit{present}, \textit{conditional}, \textit{hypothetical} and \textit{none} classes respectively.}
\label{tab:baseline_data_format}
\end{table}


\subsection{Training and Hyperparameters}
\begin{figure}[htb]
    \centering
    \includegraphics[width=\linewidth]{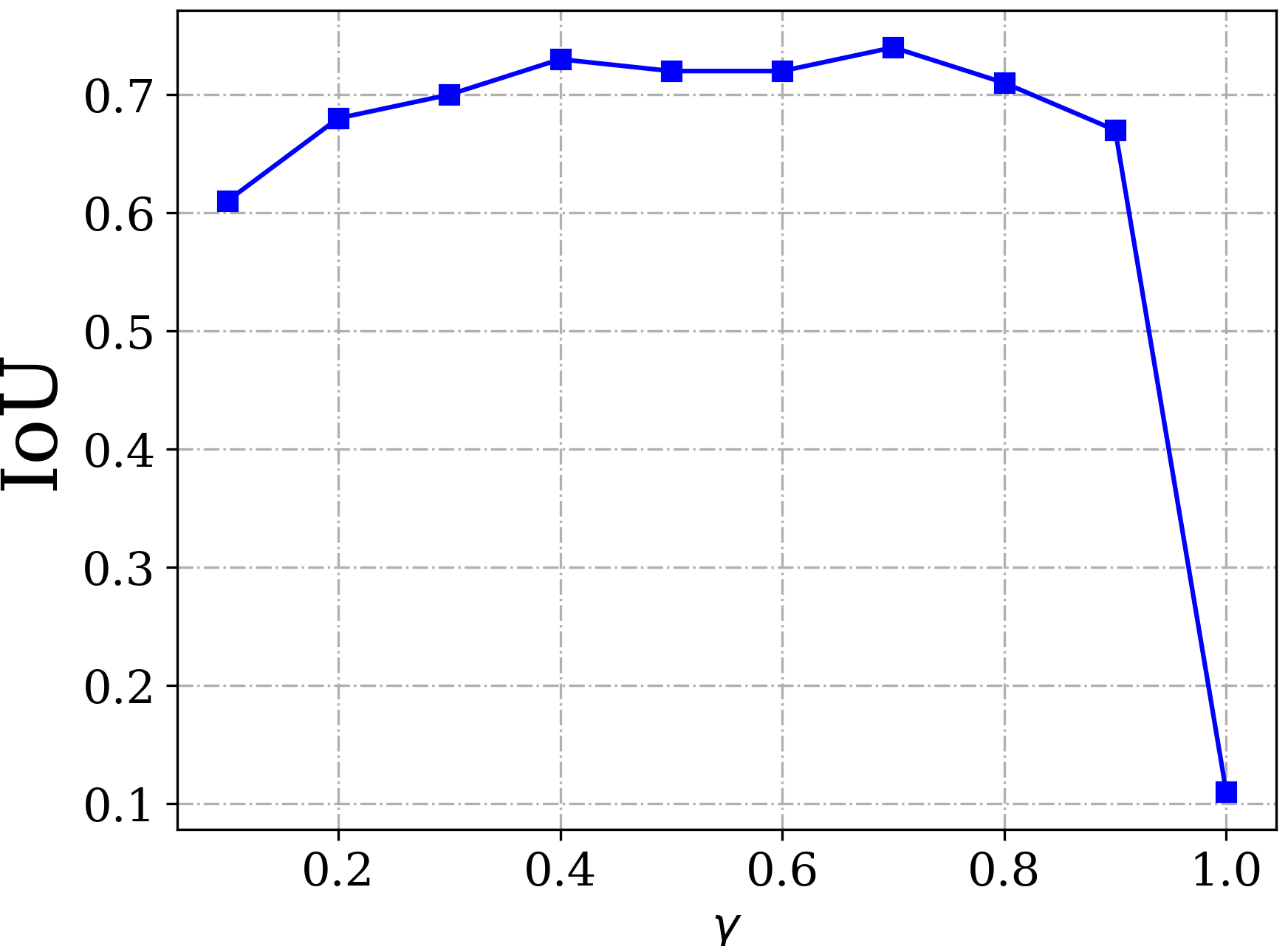}
    \caption{F1 score with different values of IOU threshold evaluated on the validation set}
    \label{fig:f1vsiou}
\end{figure}

Since the datasets have unbalanced classes, we have used stratified sampling \cite{Sechidis:2011:SMD:2034161.2034172, skmultilearn-python} to represent the classes in the same ratio in train, validation and test sets. To further mitigate the effect of unbalanced classes in each batch of training data, we weighted the cross entropy loss with  the inverse of the number of examples for each class. The pre-trained BioWord2Vec \citep{Zhang2019BioWordVecIB} is used in the embedding layer with frozen weights. We used Adam Optimizer with the default learning rate of 0.001 for 400 epochs. Shuffling after each epoch results in different distribution of classes per batch of iteration. This leads to unstable training and therefore takes more epochs for convergence. We have set the number of prior boxes to be 24, little more than the maximum length of a class scope in the training set. Fig-\ref{fig:f1vsiou} shows the performance of the model on validation set with different values of IoU threshold ($\gamma$), the maximum being $\gamma=0.7$. Experiments with more layers and higher kernel sizes didn't improve the performance. This is because the receptive field has to be large enough to span the longest scope in the input i.e 20.

\begin{figure}[ht!]
    \centering
    \includegraphics[width=\linewidth]{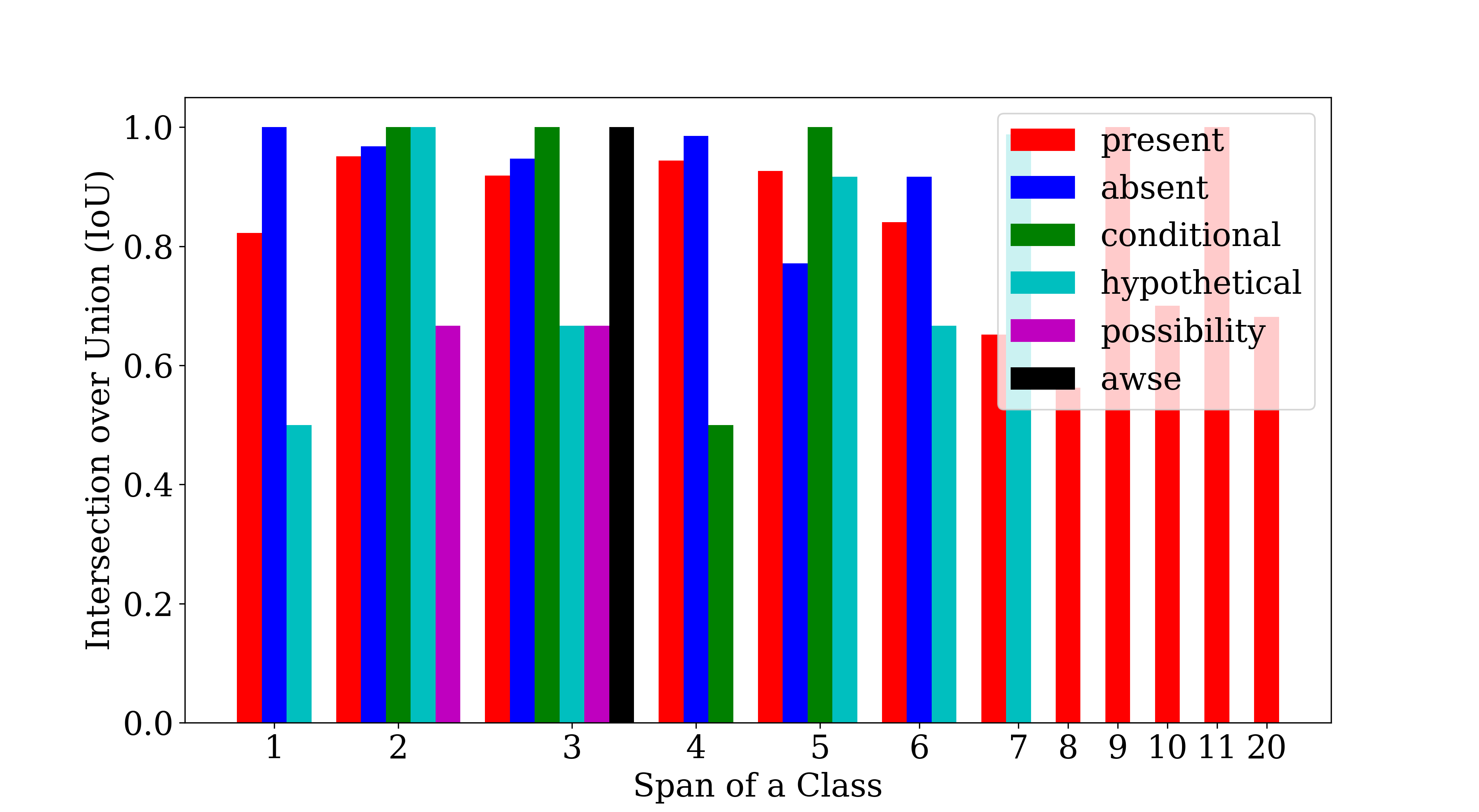}
    \caption{This figure shows the IoU (Intersection over Union) between the predictions and ground-truths on test set for different scope lengths.}
    \label{fig:results_analysis_pr}
\end{figure}

\begin{figure}[ht!]
    \centering
    \includegraphics[width=\linewidth]{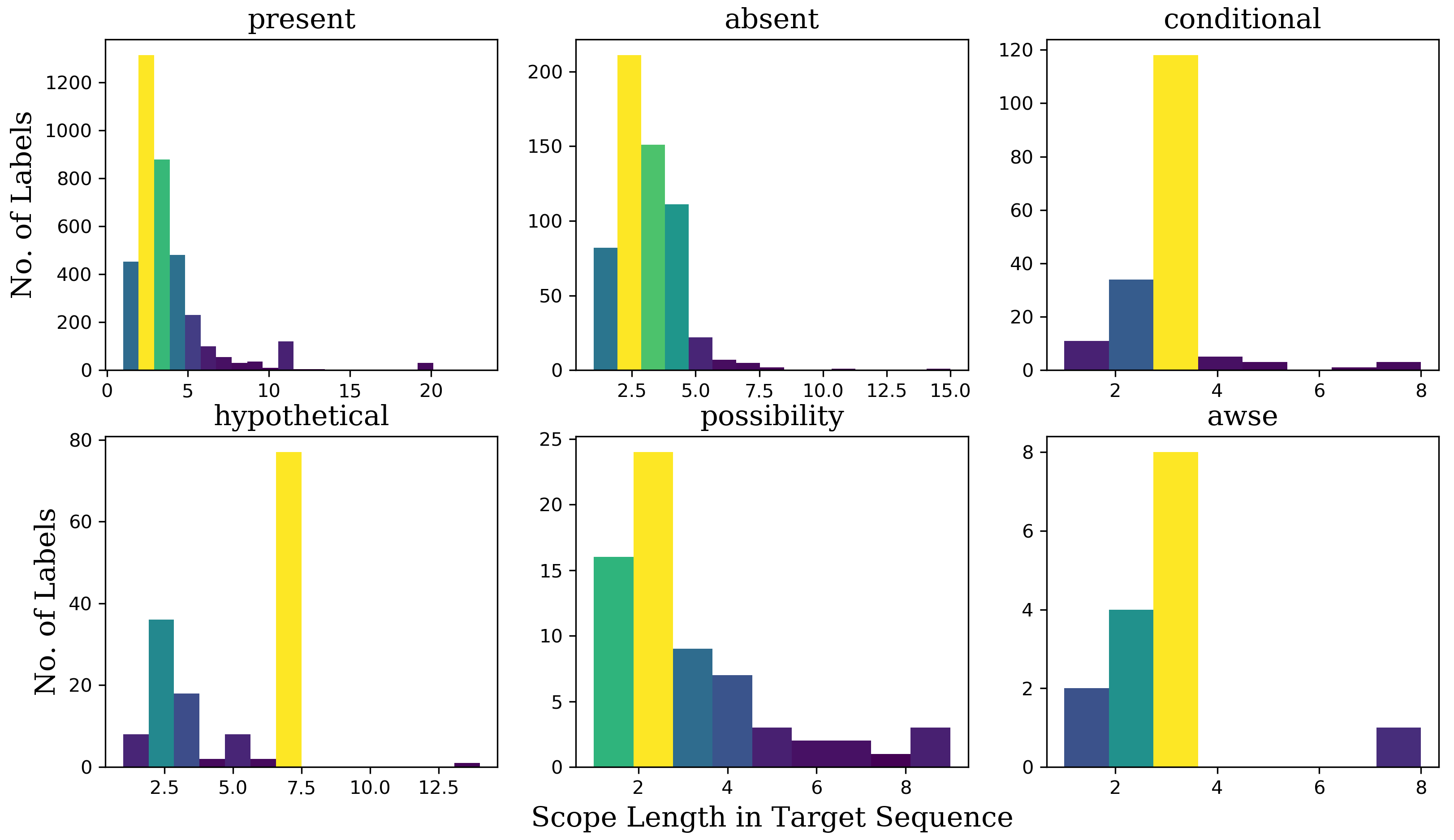}
    \caption{This figure shows histograms of scope lengths per each class in the training set.}
    \label{fig:train_span_stats}
\end{figure}

\begin{figure*}[ht!]
    \centering
    \includegraphics[width=\linewidth]{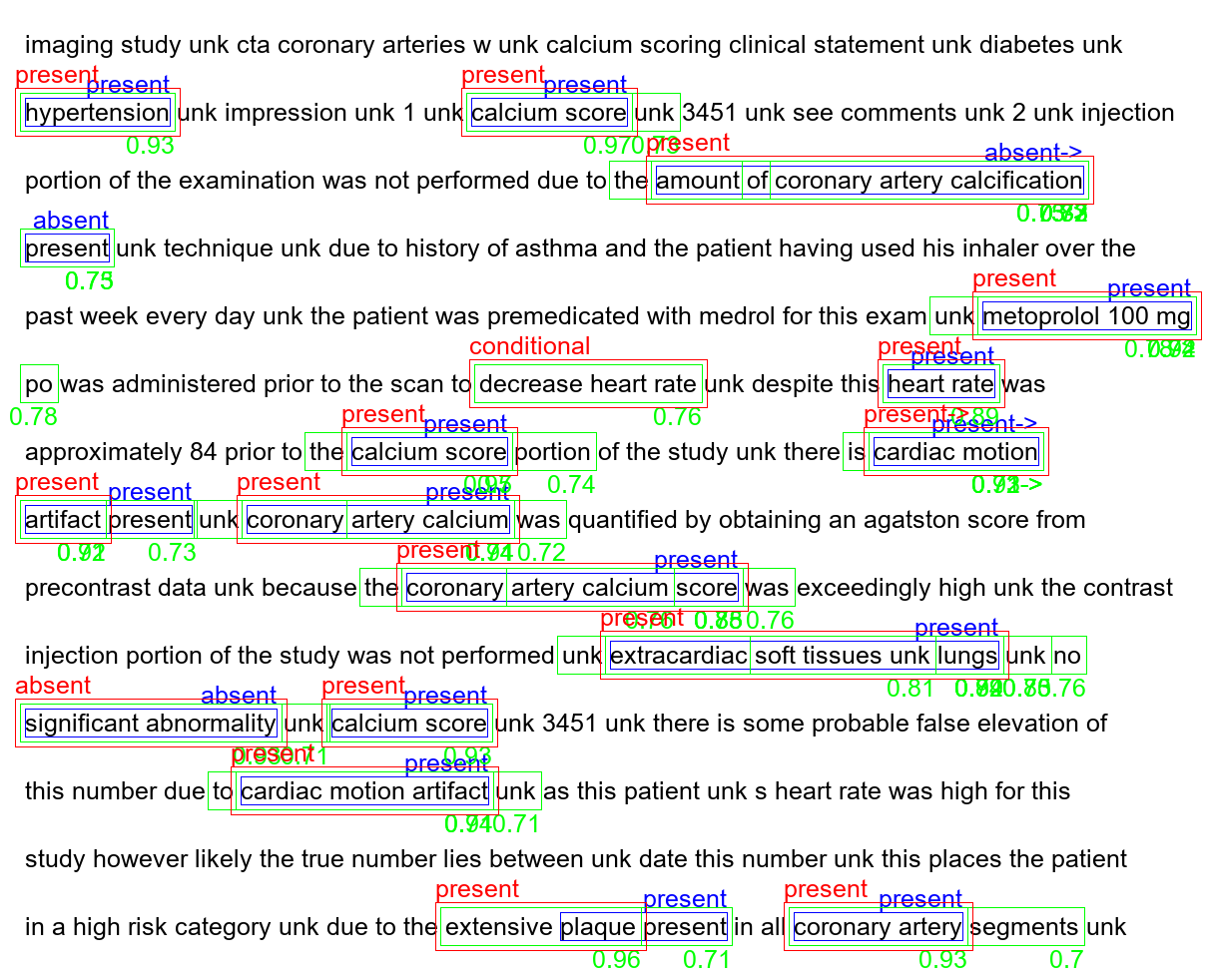}
    \caption{A sample output of the model on a report from Dataset-I. The ground-truths are shown in blue. The green boxes are prior boxes each having a box-confidence score. After non-max suppression of these prior boxes, the final predictions are shown in blue. $"->"$ is used to indicate that the box is extended to next line. "unk" represents the unknown token, it is used to represent the words that are not found in the vocabulary.}
    \label{fig:sample_output}
\end{figure*}

\subsection{Results}
Table.\ref{tab:results} shows the performance of the baseline and our CNN-based scope localization models on Datasets-I,II per each class. For a fair comparison with the baseline, the box predictions from our model are converted to a sequence of labels per token. On first impressions, the performance seem to be affected by the quantity of data available for training with the best performance on \textit{present} class and least performance on \textit{AWSE} class. After further analysis, it appears that the scope lengths found in the training set is also a crucial factor. Fig.\ref{fig:train_span_stats} shows a histogram of scope lengths available in the training set for each class. The performance on the test set for different scope lengths is shown in Fig.\ref{fig:results_analysis_pr}. As shown, model performance for the \textit{present} class declines with scope lengths 7, 10, and 20, which reflect sparsity of this class for these scopes in the training set. In contrast, the model performs well on the \textit{hypothetical} class  with scope length 7, reflective of the better distribution of this class for this scope relative to other scopes. 

\section{Conclusion}
In this work, we have explored a novel approach of scope localization and classification with a single end-to-end CNN model. We demonstrated good performance and thereby make a case for using multi-label clinical text that is often found in real world. For future work, we would like to explore the usage of inception layers; different sets of kernel sizes in each layer. The output layer will then have varying receptive fields i.e scope lengths in our problem. This increases the generalization of the model to scope lengths that are unseen in the training data.

\bibliography{ref}
\bibliographystyle{acl_natbib}

\end{document}